\documentclass{elsarticle}
\usepackage{color}
\usepackage{array}
\usepackage{textcomp}
\usepackage{multirow}
\usepackage{amsmath}
\usepackage{amssymb}
\usepackage{graphicx}
\usepackage{subfigure}
\usepackage{longtable}
\usepackage{adjustbox}
\usepackage{amsthm}
\usepackage{soul}
\usepackage{paralist}
\usepackage{eurosym}
\usepackage{float}
\usepackage[margin=2cm]{geometry}
\usepackage{caption}

\usepackage[hyphens]{url}
\usepackage{hyperref}
\hypersetup{breaklinks=true}

\usepackage{tikz,collcell}
\newcommand*{\cellset}{\pgfqkeys{/cell}}
\newcommand*{\myCell}{\cellset}
\newcolumntype{C}[1]{>{\collectcell\myCell}#1<{\endcollectcell}}
\makeatletter
\tikzset{overlay linewidth/.code=\tikz@addmode{\tikzset{overlay}}}
\cellset{.unknown/.code={\edef\pgfkeys@temp{\noexpand\cellset{box=\pgfkeyscurrentname}}\pgfkeys@temp}}
\makeatother
\cellset{
  head/.code=\rlap{\rotatebox{30}{#1}},
  myCell/.style={
    draw=black,
    fill={#1},
    text width =1.5cm,
    minimum size= 0.75cm},
  box/.code={%
    \tikz[baseline=0ex]
      \node[/cell/myCell={#1}]{};},
  box/.default=none,
  ./.style={box},
  @define/.style args={#1:#2}{#1/.style={box=#2}},
  @define/.list={g:green, y:yellow, r:red, o:orange, y:yellow}
  }

\hypersetup{pdfauthor={Name}}

\journal{arXiv}

\theoremstyle{definition}

\begin{document}

\title{Assessing Climate Transition Risks in the Colombian Processed Food Sector: A Fuzzy Logic and  Multicriteria Decision-Making Approach}

\author[l1]{Juan F. Pérez-Pérez}\ead{juan.perez50@eia.edu.co}
\author[l1]{Pablo Isaza Gómez}\ead{pablo.isaza@eia.edu.co}
\author[l1]{Isis Bonet}\ead{isis.bonet@eia.edu.co}
\author[l2]{María Solange Sánchez-Pinzón}\ead{mssanchez@serviciosnutresa.com}
\author[l3]{Fabio Caraffini\corref{cor1}}\ead{fabio.caraffini@swansea.ac.uk}
\author[l1]{Christian Lochmuller}\ead{christian.lochmuller@eia.edu.co}
\cortext[cor1]{Corresponding author}
\address[l1]{Computational Intelligence and Automation Research Group, EIA University, Envigado, Colombia}
\address[l2]{Vicepresidency of Sustainability, Grupo Nutresa, Colombia}
\address[l3]{Department of Computer Science, Swansea University, UK}

\begin{abstract}
Climate risk assessment is becoming increasingly important. For organisations, identifying and assessing climate-related risks is challenging, as they can come from multiple sources. This study identifies and assesses the main climate transition risks in the colombian processed food sector. As transition risks are vague, our approach uses Fuzzy Logic and compares it to various multi-criteria decision-making methods to classify the different climate transition risks an organisation may be exposed to. This approach allows us to use linguistic expressions for risk analysis and to better describe risks and their consequences. The results show that the risks ranked as the most critical for this organisation in their order were price volatility and raw materials availability, the change to less carbon-intensive production or consumption patterns, the increase in carbon taxes and technological change, and the associated development or implementation costs. These risks show a critical risk level, which implies that they are the most significant risks for the organisation in the case study. These results highlight the importance of investments needed to meet regulatory requirements, which are the main drivers for organisations at the financial level. 
\end{abstract}
\maketitle

\section{Introduction}\label{sec:intro}
Climate-related risks have wide-reaching implications, causing economic, material, environmental, and social damage. To mitigate these losses, data-driven models that analyse historical data are emerging, helping to improve the assessment of these risks and achieve a more accurate and comprehensive understanding of the potential impacts associated with climate-related hazards \cite{Bronnimann2019}. By leveraging the insights gained from the analysis of historical data, companies aim to improve the ability to anticipate, prepare for, and respond to the challenges posed by climate-related risks. There are several methods to assess climate risk and make decisions accordingly, and the most appropriate approach greatly depends on factors such as data requirements, time frame, and organisational purpose \cite{Hanski2019}.

Increasingly severe weather conditions pose growing systemic risks to businesses in the global economy \cite{Huang2018}. These include the impacts of climate-related phenomena on natural ecosystems, including rising sea levels, increased average temperature, changes in precipitation patterns, and temperature fluctuations. These physical effects generate significant risks for companies, commonly referred to as climate physical risks, and can lead to financial losses due to asset damage or business interruptions \cite{Sakhel2017}. In turn, the growing recognition of climate change has led to the implementation of regulatory measures to address its impact, resulting in an increasing number of climate regulations that pose financial risks for organisations, often resulting in additional operating and investment costs. Furthermore, market risks arise from changes in consumer and investor behaviour influenced by climate change. These shifts in demand and supply dynamics can have significant effects on company revenues \cite{Sakhel2017}; this phenomenon is called climate transition risks. These types of risks are classified as: regulatory, technological, market and reputational risk.

The correlation between physical risk and transition risk is clear; as transition policies are implemented, the impact of physical risks is generally mitigated. Furthermore, when physical risks are more likely to manifest, the need for effective transition measures increases \cite{Oguntuase2020}.

Climate change has become one of the most pressing global issues. It is essential to analyse the science of climate change and the uncertainties associated with it, as well as to identify and explain the associated climate change risks and the financial implications \cite{Oguntuase2020}.

Predicting the future has always been a human desire, but when it comes to climate change, its impacts, mitigation, and adaptation, there is a great deal of uncertainty. Scientists have estimated the probability of different outcomes for greenhouse gas emissions (GHG) and temperatures, but this still carries some risk.

The exact nature, timing, frequency, intensity, and location of the effects of climate change cannot be predicted due to the uncertainties involved. These uncertainties are based on a variety of demographic and socioeconomic elements, such as technology, values and preferences, and policies \cite{Kalra2014}. In addition, our limited understanding of the climate system leads to scientific uncertainty \cite{Heal2013}.

Companies must recognise, quantify, monitor, control, and communicate their exposure and vulnerability to these risks to their stakeholders. Companies should demonstrate how they are mitigating significant risks and have in place reliable policies or regulations to manage these risks.

In this situation, according to \cite{Bankinternational2021} a successful risk management system should have three goals: first, to identify the main sources of climate risk and how it is spread; second, to map and assess climate-related exposures and any areas of risk concentration; and third, to create financial risk metrics for climate risks. In this investigation, we will address the first two points; the third point will be addressed in future research.

However, a primary task would be the timely identification of climate risks for an organisation. This is articulated with the most recent regulation implemented in Colombia, the External Circular 031 of the Superintendencia Financiera (SFC), Colombia's financial regulator, which requires listed companies that are to identify the risks related to climate change to which they are exposed and to disclose them to their stakeholders.

To address this complexity, a comprehensive and integrative model is needed to weigh these factors in an interconnected way. Multicriteria Decision Making (MCDM) methods provide a structured and analytical approach to this problem, allowing for the consideration of multiple variables and the evaluation of various scenarios that could affect an organisation's risk profile. MCDM thus offers a more informed and strategic decision-making framework in the ever-changing climate context.

Companies often turn to risk matrices to make decisions, which are intended to assess the probability and effect of different risks. However, traditional approaches are inadequate to accurately model certain climate risks due to their unpredictable and ever-changing nature \cite{Kalra2014}. To address this, tools that are not only easy to update but also promote effective communication within the organisation are needed. Here, fuzzy logic (FL) can be beneficial. FL enables the consideration of multiple factors and relationships in a flexible manner and is capable of dealing with the uncertainty associated with risk assessment. Furthermore, it is well suited to the need for adaptability and communication effectiveness, making it a powerful tool for risk assessment in the unpredictable climate environment.

Therefore, this research will focus on identifying and measuring the risks of climate transition that can affect a company in the processed food sector in Colombia; this will be done by constructing a risk matrix where the probability of occurrence and impact will be evaluated, analysed from a financial perspective in the organisation, as well as the level of risk exposure, using computational techniques and expert judgment. To achieve this objective, 10 MCDM methods and the FL approach will be compared, and the one that best fits the analysed problem will be selected. We selected a fuzzy method, because FL can work with uncertainty and imprecision and solve problems where there are no defined limits or precise values \cite{Markowski2008}. This situation occurs in the concept of a risk assessment matrix, adding the uncertainty of climate change.

In this context, we are introducing a comprehensive system for identifying and addressing climate transition risks, with the aim of mitigating their potentially disruptive effects on business. Our system's main objective is to provide organisations with a robust framework that streamlines the identification and assessment of these risks and enables them to proactively implement measures that effectively minimise their impact.

The remainder of this article is structured as follows: 
\begin{itemize}
\item Section \ref{sec:literature} reviews and comments on relevant studies in the literature;
\item Section \ref{sec:materialAndMethods} describes the data and the methods employed in this research and provides some background information to better understand and motivate our research approach;
\item Section \ref{sec:fuzzySystem} presents the proposed fuzzy system;
\item Section \ref{sec:results} presents and comments on the obtained results;
    \item Section \ref{sec:conclusions} draws the conclusion of this study.
\end{itemize}

\section{Literature review}\label{sec:literature}

\subsection{Fuzzy systems in the risk assessment domain}

Fuzzy Logic (FL) is a widely used paradigm for modelling and assessing risks in various domains \cite{shapiro2015risk,pena2021fuzzy}. While modern Artificial Intelligence (AI) methods such as convolutional neural networks and decision-making frameworks are gaining popularity for risk assessment, see, e.g. \cite{pena2023multispectral}, FL still holds a crucial role in modelling risk variables and their probabilities. As a result, there are numerous studies in the current literature that combine FL with AI techniques. The most effective hybrid systems can be categorised into four groups:
\begin{enumerate}
    \item FL combined with multi-criteria decision-making methods (MCDM);
    \item FL and neural networks;
    \item FL and Bayesian Networks (BN);
    \item FL integrated with quantitative risk assessment (QRA). 
\end{enumerate}

Each category has its own advantages and disadvantages, making it more appropriate for specific contexts. 

Neurofuzzy logic systems have shown promise in improving the decision-making process and generating meaningful risk classifications \cite{Ak2019}. For example, in a study conducted in \cite{Nabipour2020}, a 5-layer adaptive neurofuzzy inference system (ANFIS) with two input variables is used to predict wind power and its variability under future climate change scenarios in the northern Caspian Sea. The system employs four Gaussian membership functions for data processing in the first stage with good classification results. Similarly, studies in \cite{Gocić2015} and \cite{Bacanli2009} apply multiple ANFIS systems to predict drought and estimate a quantitative measure of drought in the form of the standardised precipitation index on time scales from 1 to 12 months. The results are satisfactory and show high accuracy in the prediction of drought.

However, due to their flexibility, MCDM methods have been widely used in conjunction with fuzzy logic, especially in risk management. These approaches provide decision makers with the means to overcome the limitations of conventional risk prioritisation methods, making them valuable tools in the field \cite{Karasan2021}. The relevant work is given in \cite{Djenadic2022} and \cite{AkGul2019}. These use the ``analytical hierarchy process'' method to define the mutual impact of partial indicators (occurrence, severity, and detectability) on the risk. To enhance the accuracy of the results, the fuzzy model incorporates the Technique of Preference Order in Similarity to the Ideal Solution (TOPSIS), which helps minimise the dispersion of the results. By extending this approach to fuzzy sets, the model further improves the accuracy and reliability of the results. An interesting review of the literature focused on the application of TOPSIS for provider selection is presented in \cite{Hajiaghaei-Keshteli2023}. In this study, three experts evaluated five alternative TOPIS-based systems using fuzzy logic linguistic expressions. Similarly, \cite{Beiranvand2023}[26] uses a fuzzy decision matrix to assess options based on expert ratings based on various criteria, using TOPSIS as the MCDM method. Failure Modes and Effects Analysis (FMEA) \cite{Karasan2021} is another commonly used MCDM method for risk analysis. However, it is important to note that FMEA has certain limitations. For example, it can produce the same level of risk for different combinations of risk variable scores, which can lead to misleading results in practical risk analysis. Furthermore, FMEA assigns equal weights to each risk parameter, which can be a disadvantage when determining the criteria weights. To address these limitations, researchers have integrated fuzzy logic and TOPSIS with the classical FMEA method to provide a more comprehensive and effective approach to risk assessment.

The combination of FL and BN has been explored in research studies such as \cite{Aydin2021} and \cite{Senol2021} to address the limitations of risk assessment, as BNs offer a compact representation of joint probability distributions, while FL provides a means to handle vagueness and ambiguity in probabilistic risk assessment. Assigning appropriate probabilities to nodes in the BN architecture is crucial to obtaining meaningful results. By incorporating fuzzy sets, the Fuzzy Bayesian Network (FBN) approach enables the modelling and management of uncertainty by utilising linguistic values in the decision-making process. A FBN is particularly useful when there is little or insufficient data to determine the probability values of the nodes in the network.

Finally, in the domain of quantitative risk assessment (QRA), a combination of fuzzy logic and quantitative methods has been explored. For example, in the context of the oil industry, \cite{Seddik2023} proposed a fuzzy QRA approach based on fuzzy logic. This approach allows for the analysis, evaluation, and mitigation of key risks associated with the industry. The models developed in this study consider factors such as the frequency and severity of consequences, as well as the level of individual and social risk. The authors used a multistep methodology the includes the calculation of the fuzzy frequency of consequences and the individual risk associated with each variable.

These risk assessment systems have been used in several application domains, such as construction \cite{Abed2023,Biswas2019,Chaher2016}, mining industry \cite{Tripathy2018,Ak2019,Aliyev2022}, health sector \cite{Casalino2020}, manufacturing sector \cite{Studenikin2021,Vahdani2015,Gallab2019,Seker2021}, and climate physical risks \cite{Pushpa2023,Gocić2015,Bacanli2009,Katipoğlu2022,Zamelina2022}. However, currently there is no significant study using these approaches in the context of transition risks associated with climate change.

\subsection{Multicriteria decision-making (MCDM) methods }

The purpose of MCDM methods is to generate a ranking of alternatives or variables considering the evaluation of multiple criteria simultaneously. However, decision-making becomes complex when there are several criteria, so various methods have been developed to facilitate this process \cite{Aydin2022}. These methods have been widely studied and have been applied to various problems to facilitate decision-making.

When analysing a specific problem, there is no standard process for selecting a method and employing it for the solution. This dilemma has been discussed in the literature because, depending on the applied MCDM method, the results may be different \cite{Ceballos2016}. For the above reasons, we will present in general terms several methods widely used by experts and then compare the results in constructing a climate transition risk ranking.

TOPSIS is a method that finds the best alternatives by measuring the distance between the chosen alternatives and the ideal solutions (the ideal solution refers to the phenomenon being evaluated). It is based on the concept that the chosen alternatives should have the shortest distance from the positive ideal solution and the longest distance from the negative ideal solution. This means that the best alternatives are those that are closest to the positive ideal solution and farthest from the negative ideal solution \cite{Wu2016}.

The COPRAS method was developed in 1996 by Zavadskas and Kaklauskas \cite{Zavadskas2009}. Like TOPSIS, this method selects the most appropriate alternative considering the best and worst ideal solutions \cite{Nilay2020}.

The BORDA counting method was developed by Jean-Charles de Borda in 1784. It is a procedure that aims to classify the alternatives according to the sum of the individual preferences of the decision makers. This is a data aggregation technique that reduces two or more classification formats to a more rational one. The method is quite easy to apply \cite{Topcu2019}.

The simple additive weighting (SAW) method is one of the most widely used additive weighting methods for aggregating a decision maker's preferences defined by multiple criteria. In this technique, the criteria are mutually independent. Preferences are based on the performance score of an alternative for each criterion, which is weighted by the weight of the respective criterion. The decision maker defines the weight of the criterion. The overall evaluation score of an alternative is calculated by summing the weighted score of that alternative according to several criteria \cite{Kumar2022}.

The ELECTRE method involves a comparison of alternatives based on how their evaluations and preference weights correspond to the pairwise dominance between them. It looks at both the agreement between the preference weights and the dominance relationships, as well as the difference between the weighted scores. This process is known as "concordance and discordance" and also as known as concordance analysis \cite{Odu2019}. The concordance and discordance measure the satisfaction and dissatisfaction from the decision maker's perspective. The threshold concept is then used to create a core of preferred alternatives \cite{Jafaryeganeh2020}. Thus, ELECTRE is an "outranking" method.

The VIKOR technique was developed by Serafim Opricovic to solve decision-making issues with conflicting and incommensurable criteria, with the understanding that compromise is an acceptable way to resolve conflicts. This approach focuses on ranking and selecting from a group of options and determines the compromise solution that is closest to the optimal solution \cite{Odu2019}.

In 2020, Stevic introduced the MARCOS method for supplier selection in the healthcare industry. This method takes a novel approach to problem solving by considering both antiideal and ideal solutions in the early stages. It also proposes a new way of calculating utility functions and combining them. This should help to ensure stability when dealing with a large number of alternatives and criteria \cite{Paradowski2022}.

The PROMETHEE approach is used for the top-level ranking. This method uses a function that reflects the degree of benefit of one option over another, as well as the degree of disadvantage. PROMETHEE involves a mutual comparison of each pair of alternatives with respect to the given criteria. PROMETHEE I is designed to generate a partial ranking. Partial rankings concentrate on the best choice, not on a full ranking. The PROMETHEE II method produces a full ranking, from the best to the worst alternative \cite{Vries2020}.

The Weighted Sum Method is a popular choice among WSM techniques, as it is easy to use and follows a logical process. The basic assumption of WSM is that the attributes are independent of each other, which means that the contribution of each attribute to the overall score is not affected by the values of the other attributes \cite{Jafaryeganeh2020}.

Finally, the CODAS method was first proposed by Ghorabaee et al. 2016, this decision-making technique uses two distance measures, the Euclidean distance and the Hamming distance, to assess the desirability of alternatives by measuring the distance from the ideal negative solution \cite{Ghorabaee2016}.

\section{Materials and methods}\label{sec:materialAndMethods}

In the field of organisational risk management, the first step involves assessing the potential risk of sources. In the specific context of climate transition risks, this process consists of first identifying the different risks and then prioritising them. The latter activity is particularly important because it allows the organisation to lay a foundation to understand potential challenges, gain valuable information about various sources and levels of risk associated with climate change, make more effective decisions, and allocate resources to mitigation and adaptation strategies.

In this study, we identified several risks by reviewing the literature on the most important risks for organisations in the processed food sector in Colombia. To validate our findings, we used expert judgement from an organisation in this sector, which allowed a better assessment and understanding of the variables selected previously.

\subsection{Risk matrix construction}

\begin{figure}[htbp]
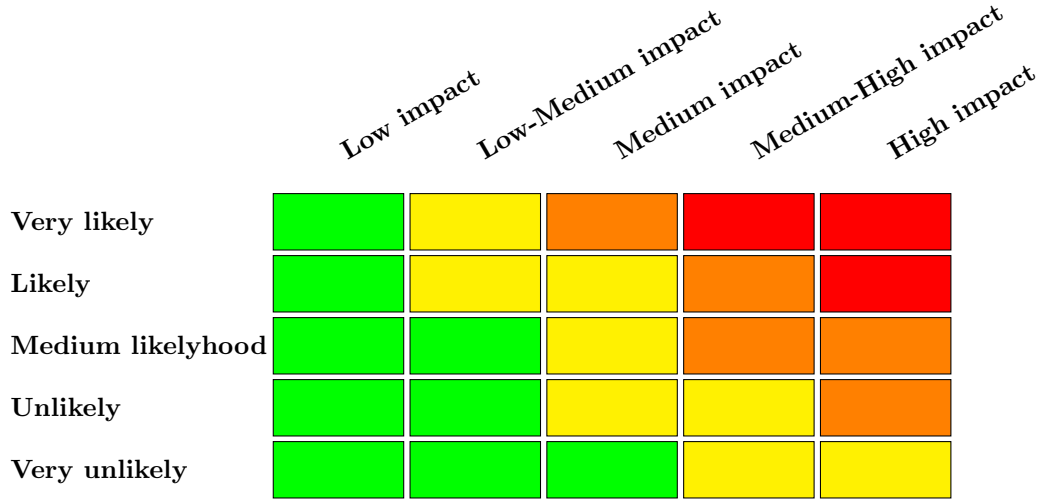

\centering
\setlength{\tabcolsep}{1pt}
\begin{tabular}{>{\footnotesize}r*5{C{c}}}
    & head=\textbf{Low impact} & head=\textbf{Low-Medium impact} & head=\textbf{Medium impact} & head=\textbf{Medium-High impact\,} & head=\textbf{High impact} \\[0.9em]
    \multicolumn{1}{l}{\multirow{1.5}{*}{\textbf{Very likely}}} &   g &   y  &  o  &    r  & r \\[0.9em]
\multicolumn{1}{l}{\multirow{1.5}{*}{\textbf{Likely}}} & g &  y  &   y & o   &   r\\[0.9em]
  \multicolumn{1}{l}{\multirow{1.5}{*}{\textbf{Medium likelyhood}}} & g & g    & y    & o       & o  \\[0.9em]
   \multicolumn{1}{l}{\multirow{1.5}{*}{\textbf{Unlikely}}} &  g & g     & y             & y&  o \\[0.9em]
    \multicolumn{1}{l}{\multirow{1.5}{*}{\textbf{Very unlikely}}} &  g & g  & g    & y   & y  
\end{tabular}\\
\caption{Risk matrix example. {\textbf{Green}} is used to indicate \textit{low risk}, {\textbf{yellow}} to indicate \textit{medium risk}, {\textbf{orange}} to indicate \textit{high risk} and {\textbf{red}} to indicate \textit{critical risk}.}
\label{fig:matrix}
\end{figure}

An example of a risk matrix is shown in Figure \ref{fig:matrix}. This visual representation allows a rapid assessment of risks by providing an intuitive way to gauge their likelihood and the resulting impact with the naked eye. It is important that the risk matrix shows the organisation's appetite for climate change risk. It is also easy to create and read even without in-depth technical or domain expertise. For these reasons, the risk matrix is a popular and widely used tool in organisations to determine which risks are more important and require more attention. Note that this method is based on a subjective definition of risk that combines the severity of the consequences and the expected frequency of the event \cite{Zaky2018}.

\subsection{Variables selection (climate transition risk identification)}

on in organisations in the processed food sector.
The initial step in developing a risk management plan is to classify the sources of risk within an organisation. To do this, we conducted an extensive review of the literature to identify the major risks of climate change that affect the processed food industry. This enabled us to identify the relevant variables to be considered in our system and rank them according to their severity. Risks were identified for the four types of risks that make up the climate transition risk, which are: regulatory, technological, market, and reputational risk.

\subsection{Multicriteria decision making methods and their comparison}

This study considers a varied set of multiple state-of-the-art MCDM methods that have very different logic and combines them with fuzzy logic to construct a climate transition risk matrix. These methods are listed below.
\begin{itemize}
    \item Technique for Order Preference by Similarity to Ideal Solution (TOPSIS) \cite{Wu2016}.
    \item Complex Proportional Assessment (COPRAS) \cite{Zavadskas2009}.
    \item Borda count method (BORDA) \cite{Topcu2019}.
    \item Simple Additive Weighting (SAW)\cite{Kumar2022}.
    \item ÉLimination Et Choix Traduisant la REalité (ELECTRE) \cite{Odu2019}.
    \item Vice Kriterijumsa Optimizacija I Kompromisno Resenje (VIKOR) \cite{Odu2019}.
    \item Measurement Alternatives and Ranking according to the Compromise Solution (MARCOS) \cite{Paradowski2022}.
    \item Preference Ranking Organization METHod for Enrichment Evaluations (PROMETHEE) \cite{Vries2020}.
    \item Weighted Sum Model (WSM) \cite{Jafaryeganeh2020}.
    \item COmbinative Distance-Based Assessment (CODAS) \cite{Ghorabaee2016}.
\end{itemize}

The Python \textit{pyDecision} library is used to allow a fair and extensive comparison of the different methods and variables considered. This comparative analysis will include established approaches from the literature, where the weights used in the risk assessment process are traditionally obtained with an FL-based system and our innovative approach that uses TOPIS to obtain a weight for each risk variable. Using \textit{pyDecision}, we can rank each of the methods under consideration (that is, the Kendall rank correlation coefficient \cite{Abdelkader2023} is used to measure the association with the number of concordances and discordances in the paired observations) and select the most appropriate one for the case at hand. 

This analysis of risk prioritisation systems enables us to evaluate the current state of the field and comprehend how far along it is based on the judgments and views expressed by experts in the literature.

\subsection{Data collection}\label{sec:datacollection}

A validation questionnaire was designed to collect feedback from experts from a Colombian organisation\footnote{The organisation providing us with feedback and data decided to remain anonymous.} in the sector. The questionnaire consists of six sections, and a facsimile version can be viewed and taken from \cite{bib:collectionForm}. The individuals who took the questionnaire were selected to have experience and areas-specific knowledge associated with this type of risk.

\subsubsection{Data set preparation}

The real data, i.e., in the form obtained from this collection phase, could not be directly used and disclosed in this article to avoid leaks of confidential information, as previously agreed with the organisation supporting this research. 

Hence, the real data set was modelled by deriving the corresponding distributions to be used to generate all the data for this study instead. The latter imitates the original data set, being equally distributed, and it is not subject to any non-disclosure agreement. We discussed this with our business partners, who agreed to adopt this approach.

\subsubsection{Design considerations on the data collection form}

The first section of the questionnaire is designed to evaluate the five criteria chosen to analyse the risk variables: vulnerability, resilience, exposure, probability, and impact. These are key to risk assessment in organisations and should all be considered when performing a detailed analysis that aims at a high-quality assessment. In fact, it is vital to consider probability, noting that risk is not limited to hazard and exposure alone, as there are other important aspects to consider, such as vulnerability, resilience, and financial impact \cite{Bingler2020}. It should be noted that the Intergovernmental Panel on Climate Change (IPCC) argues that the risks of climate disasters are composed of exposure, vulnerability, and risk. This approach is also useful for assessing the financial impacts of climate transition risks. Often, economic and financial analyses consider only the exposure of an organisation, but it is essential to consider vulnerability, resilience, and economic impact to fully understand the potential materialisation of these climate-related risks, as these concepts are interdependent. For example, if a company is less resilient than others, then its exposure may be higher; if a company is more vulnerable, then its exposure may also be higher \cite{IPCC2014}.

\begin{figure}[ht!]
\centerline{\includegraphics[width=0.8\textwidth]{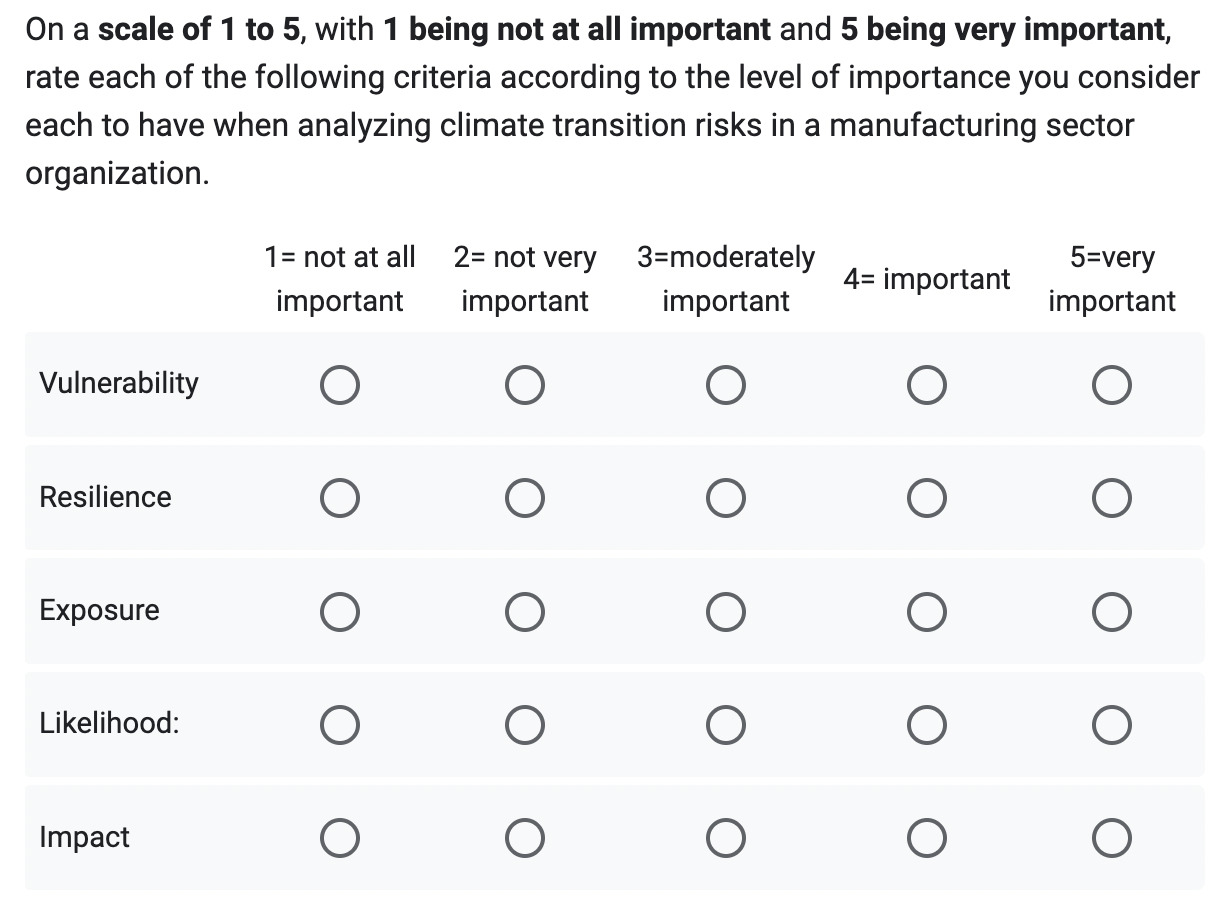}}
\caption{Section 1: transitional climate risk criteria rating.}
\label{fig:Criterion}
\end{figure}

Due to the considerations above, we decided to determine the importance of the risk assessment criteria with feedback from experts in the field. In the first section of the questionnaire, they were therefore asked to rate the criteria on a ``Likert scale'' ranging from 1 to 5, with 1 representing the minimum importance and 5 representing the maximum importance (see Figure \ref{fig:Criterion}).

\begin{figure}[ht!]
\centerline{\includegraphics[width=0.8\textwidth]{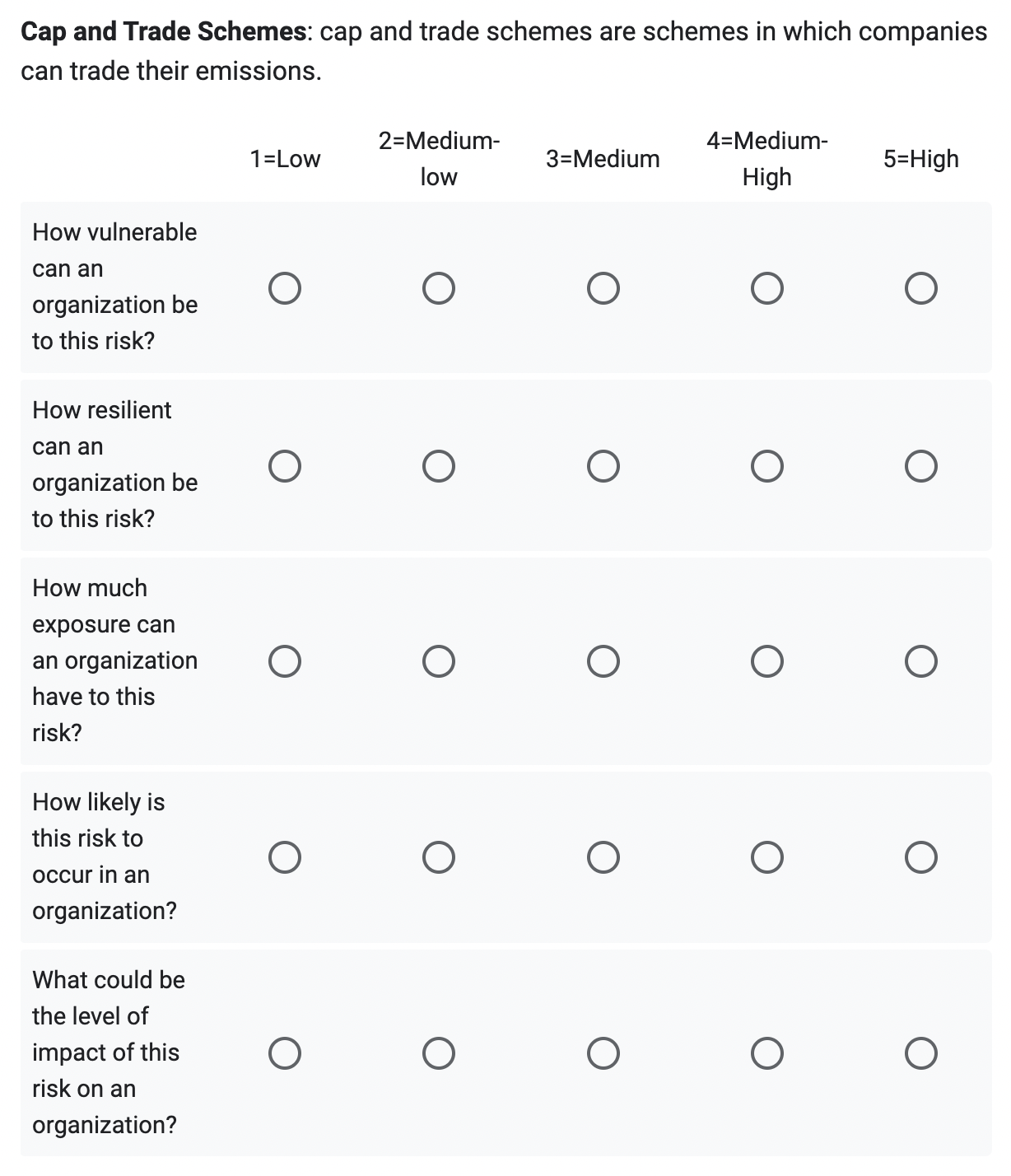}}
\caption{Section 2: regulatory risks.}
\label{fig:example_risk}
\end{figure}

In the second section of the data collection form, we introduce four regulatory climate risks (see Figure \ref{fig:example_risk}). 

In this section, assessors were asked to rate criteria on a scale of 1 to 5, with 1 being the lowest rating and 5 being the highest for each criterion. The financial effect of each risk on an organisation is reflected in the qualitative ratings. It is worth pointing out that what may be considered as carrying a high impact by one expert, another expert may consider to have a medium impact. Therefore, because of their ``fuzzy'' nature, we agreed that converting them into fuzzy variables would be the most logical approach.

Similar considerations were made for each of the four remaining variables when deciding what to include in the questionnaire form. With reference to \cite{bib:collectionForm}, one can see that variables such as market and reputation are addressed in sections 3, 4 and 5 of the proposed questionnaire.

Finally, the sixth section of the form focusses on defining the rules to use in the fuzzy logic model. In this section, experts were asked to assign the level of risk (low, medium, high, or critical) to the 25 combinations of rules introduced in Section \ref{sec:fuzzySystem} of this manuscript.

\section{The proposed fuzzy system}\label{sec:fuzzySystem}

Perception of risk and the relationship between its variables can vary between individuals. Fuzzy logic is a valuable tool in risk management and assessment because it allows the use of human expertise to create fuzzy if-then rules. It is essential to take advantage of the knowledge of risk managers and experts and convert it into rules that can be used by a fuzzy inference system (FIS) to automate risk assessment.

For the sake of reproducibility, the code that implements the FIS described in the rest of this article is made available as an attachment to the online version of this manuscript.

\subsubsection{Fuzzy risk matrix sets definition (fuzzification)}

After assessing the risks with the help of experts, we proceed to the fuzzification stage. In this stage, the data are transformed into fuzzy values by mapping the input values into fuzzy set membership functions. These fuzzy values are then processed by the reasoning system, also known as FIS \cite{Ulfah2022}.

To create a fuzzy risk matrix, it is necessary to select the relevant and available input variables and divide their range into fuzzy sets. In addition, different forms of membership functions can be used, depending on the characteristics of the variables, and this choice will depend on the applications where the fuzzy systems will be implemented. For the case of risk analysis, the normalised trapezoidal membership function type is generally used, which is a commonly used option to quantify the certainty of expert opinions. These membership functions are represented by a tuple (a1, a2, a3, a4), where ${a1 \le a2 \le a3 \le a4}$, and have a height of 1. This choice is considered the most natural for this type of system \cite{Cestero2013}.

\subsubsection{Fuzzy inference system}

An FIS uses rules based on risk knowledge to map fuzzy input sets (frequency and severity) to fuzzy output risk sets. This is accomplished by using fuzzy IF-THEN rules. The fuzzy rule structure for the fuzzy risk matrix is as follow: IF the probability is ${\overline{p}_n}$ AND the impact of the consequences is ${\overline{i}_n}$ THEN the risk level is ${\overline{r}_n}$, where ${\overline{p}_n}$, ${\overline{i}_n}$, and ${\overline{r}_n}$ are the fuzzy sets for the probability, impact and risk level defined over the discourse universes, respectively \cite{Markowski2008}.

In this case, five levels of probability and five levels of impact are used, generating a total of 25 rules representing the risk categories. These fuzzy rules capture the knowledge and experience of experts through linguistic variables and are combined with data analysis to obtain more accurate and realistic results (see Table \ref{tab:table_1}).

\begin{table}[ht!]
    \centering
    \begin{adjustbox}{width = \linewidth}
    \begin{tabular}{c|l}
    \hline\hline
\textbf{Rule} & \textbf{Description} \\ 
\hline
1 & IF the likelihood is LOW and the impact is LOW, THEN the risk level IS LOW \\
2 & IF the likelihood is LOW and the impact is LOW-MEDIUM, THEN the risk level IS LOW \\
3 & IF the likelihood is LOW and the impact is MEDIUM, THEN the risk level IS LOW \\
4 & IF the likelihood is LOW and the impact is MEDIUM-HIGH, THEN the risk level IS MEDIUM \\
5 & IF the likelihood is LOW and the impact is HIGH, THEN the risk level IS MEDIUM \\ 
6 & IF the likelihood is LOW-MEDIUM and the impact is LOW, THEN the risk level IS LOW\\ 
7 & IF the likelihood is LOW-MEDIUM and the impact is LOW-MEDIUM, THEN the risk level IS LOW\\ 
8 & IF the likelihood is LOW-MEDIUM and the impact is MEDIUM, THEN the risk level IS MEDIUM\\
9 & IF the likelihood is LOW-MEDIUM and the impact is MEDIUM-HIGH, THEN the risk level IS MEDIUM\\
10 & IF the likelihood is LOW-MEDIUM and the impact is HIGH, THEN the risk level IS HIGH\\
11 & IF the likelihood is MEDIUM and the impact is LOW, THEN the risk level IS LOW\\
12 & IF the likelihood is MEDIUM and the impact is LOW-MEDIUM, THEN the risk level IS LOW\\
13 & IF the likelihood is MEDIUM and the impact is MEDIUM, THEN the risk level IS MEDIUM\\
14 & IF the likelihood is MEDIUM and the impact is MEDIUM.HIGH, THEN the risk level IS HIGH\\
15 & IF the likelihood is MEDIUM and the impact is HIGH, THEN the risk level IS HIGH\\
16 & IF the likelihood is MEDIUM-HIGH and the impact is LOW, THEN the risk level IS LOW\\
17 & IF the likelihood is MEDIUM-HIGH and the impact is LOW-MEDIUM, THEN the risk level IS MEDIUM\\
18 & IF the likelihood is MEDIUM-HIGH and the impact is MEDIUM, THEN the risk level IS MEDIUM\\
19 & IF the likelihood is MEDIUM-HIGH and the impact is MEDIUM-HIGH, THEN the risk level IS HIGH\\
20 & IF the likelihood is MEDIUM-HIGH and the impact is HIGH, THEN the risk level IS CRITICAL\\
21 & IF the likelihood is HIGH and the impact is LOW, THEN the risk level IS LOW\\
22 & IF the likelihood is HIGH and the impact is LOW-MEDIUM, THEN the risk level IS MEDIUM\\
23 & IF the likelihood is HIGH and the impact is MEDIUM, THEN the risk level IS HIGH\\
24 & IF the likelihood is HIGH and the impact is MEDIUM-HIGH, THEN the risk level IS CRITICAL\\
25 & IF the likelihood is HIGH and the impact is HIGH, THEN the risk level IS CRITICAL\\
\hline\hline
\end{tabular}
\end{adjustbox}
\caption{Fuzzy rules set}
\label{tab:table_1}
\end{table}

Mamdani's fuzzy inference algorithm is used to convert qualitative rules into quantitative results. Mamdani systems are advantageous for expert system applications due to their intuitive and straightforward rule bases, which are derived from human expert knowledge. This model employs the min operator for AND logic and the implication of the output set. The output fuzzy sets are aggregated from the evaluated rules. The aggregate membership function for an output fuzzy risk category is calculated using the maximum of the minima between the frequency sets, severity sets and risk sets (see Equation 8). The model was implemented using the ``sckit-fuzzy'' library in Python.

\begin{figure}[ht!] \centering
\subfigure[Probability]{\includegraphics[width=0.3\textwidth]{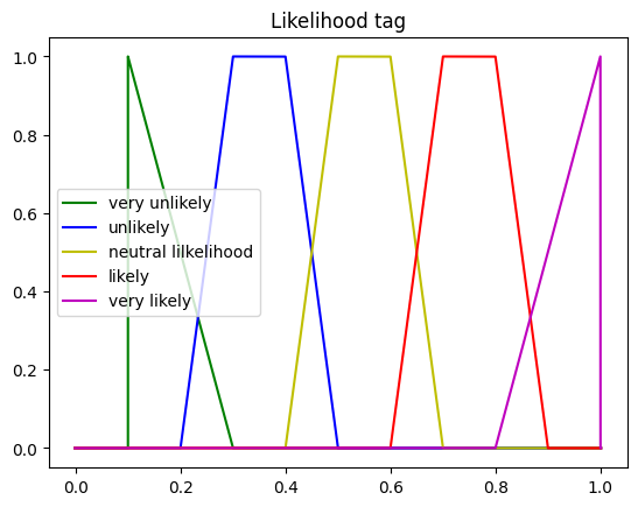}\label{fig:Fig_3}}
\subfigure[Impact]{\includegraphics[width=0.3\textwidth]{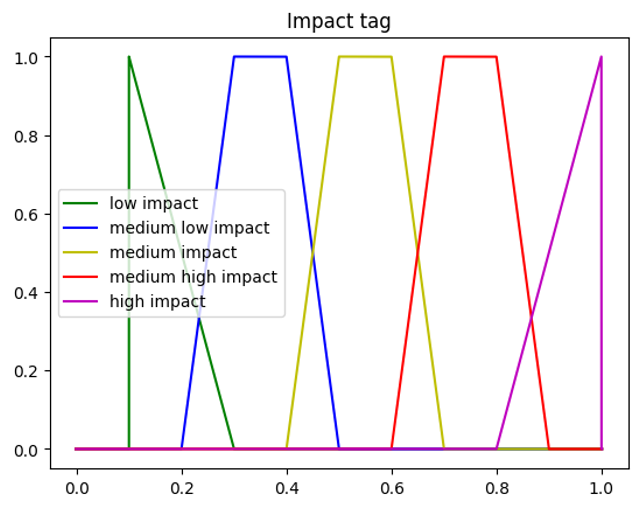}\label{fig:Fig_4}}
\subfigure[Risk level]{\includegraphics[width=0.3\textwidth]{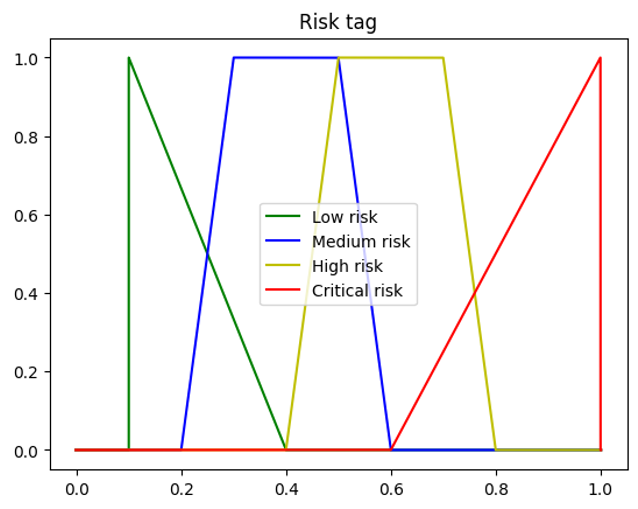}\label{fig:Fig_5}}
\caption{membership functions.}\label{fig:membershipFunctions}
\end{figure}

In addition, the graph of Figure \ref{fig:membershipFunctions} shows the membership functions. 

In particular, Figure \ref{fig:Fig_3} shows the trapezoidal membership functions of the first entry of the risk matrix, which was designed based on an FIS related to the probability of the risk, at a level of five scales: very unlikely, unlikely, neutral probability, probable and very probable.

On the other hand, Figure \ref{fig:Fig_4} shows the membership functions of the second entry of the risk matrix, with respect to the impact on the low, medium-low, medium-high and high scales.

Finally, Figure \ref{fig:Fig_5} shows the third component of the membership functions of the matrix fuzzy system model, the risk levels, which are represented on the scales low, medium, high and critical.

Table \ref{tab:table_2} summarises the classification and description of the risk levels identified by the case study organisation.

\begin{table}[ht!]
    \centering
    \begin{adjustbox}{width = 1\linewidth}
    \begin{tabular}{c|cccc}
    \hline\hline
\textbf{Risk level} & \textbf{Description} \\ 
\hline
\multirow{2}{*}{Low} & The probability that an adverse event will occur is low and the consequences are minimal or manageable. \\
& This type of risk can be easily mitigated with preventive measures. \\
\hline
\multirow{2}{*}{Medium} &The probability that an adverse event will occur is moderate with significant but not severe consequences.  \\
 & Adequate risk management is required to reduce the probability of occurrence and mitigate impacts. \\
\hline
\multirow{2}{*}{High} & The probability that an adverse event will occur is high and has significant consequences. A careful risk   \\
 & management and a proactive approach are required to minimise risk and possible negative consequences. \\
\hline
\multirow{3}{*}{Critical} & The probability that an adverse event will occur is extremely high and has catastrophic consequences. \\
& These can include both severe financial losses and a deleterious impact on the environment.\\
 &  This type of risk requires rigorous management and complete contingency planning. \\
\hline\hline
\end{tabular}
\end{adjustbox}
\caption{Risk levels}
\label{tab:table_2}
\end{table}

The inference engine of the system evaluates each rule stored in the knowledge base and performs fuzzy inference. This is achieved by determining the strength of each fuzzy rule, considering the degree of coincidence and the fuzzy connectors used in the antecedents of the rule. This allows decisions to be made in a system that considers not only true or false values, but also fuzzy values \cite{Fakhravar2020}.

The defuzzification module of the model is responsible for producing a single-scalar output or a crisp value based on the inference result. Since in the fuzzification module, the crisp values of the input variables are fuzzified into the fuzzy sets, this final stage extracts the precise value from the range of the fuzzy set for the output variable. In this case, the centroid method or the centre of the area (CAO) method, also known as a centre of the area or centre of gravity, since the centroid is the point along the x-axis around which the fuzzy set would balance \cite{Fakhravar2020}.

According to \cite{Markowski2008} the COA calculates the weighted average of a fuzzy set. The result of applying the COA defuzzification to the risk index can be expressed with Equation \ref{eq:riskInd}, which is displayed below.

\begin{equation}\label{eq:riskInd}
r = \frac{\int \mu_{\overline{R}} (r) rdr}{\int \mu_{\overline{R}} (r)rdr} 
\end{equation}

 The questionnaire form introduced in Section \ref{sec:datacollection} was used to collect feedback from experts to create our data set and produce results through our proposed system.

\section{Experimental phase and results discussion}\label{sec:results}

Our results refer to the data and methods described in Section \ref{sec:materialAndMethods}.

\subsection{Selected variables}

A critical step in managing climate transition risks in an organisation is to categorise and prioritise risk sources, allowing the allocation of the necessary resources to mitigate their impacts \cite{Ulfah2022}. In this regard, variables based on each of the four types of risks were identified during the literature review.

Table \ref{tab:table_3} shows the variables selected for expert evaluation using the questionnaire explained in Section \ref{sec:datacollection}. Risks are classified by type, including regulatory, technological, market, and reputational risks, as well as their description and the sources supporting their selection.

\begin{table}[ht!]
    \centering
    \begin{adjustbox}{ width = \linewidth}
    \begin{tabular}{c|c|c|c}
    \hline\hline
\textbf{Risk name} & \textbf{Risk type} & \textbf{Description} & \textbf{Sources}\\ 
\hline
\multirow{2}{*}{Cap and Trade Schemes} & \multirow{7}{*}{Regulatory} & Emission trading systems with pre-established caps are systems  &	\multirow{2}{*}{\cite{Bronnimann2019,Hanski2019,Huang2018}}\\
 &  & in which companies can trade their emissions &	\\
\cline{0-0}\cline{3-4}
\multirow{2}{*}{Carbon tax increase} &  & \multirow{2}{*}{Regulatory strategy to reduce greenhouse gas emissions from companies} & \cite{Bronnimann2019,Hanski2019,Huang2018,Sakhel2017,Abed2023,Biswas2019,Chaher2016,Tripathy2018,Ak2019,Aliyev2022}\\
&  &  & \cite{Casalino2020,Studenikin2021,Vahdani2015,Gallab2019,Seker2021,Pushpa2023,Gocić2015,Bacanli2009,Katipoğlu2022}\\
\cline{0-0}\cline{3-4}
\multirow{2}{*}{Climate change-related litigation} &  & Legal risks for non-compliance with climate change-related responsibilities,  & \multirow{2}{*}{\cite{Huang2018,Tripathy2018,Ak2019,Seker2021,Zamelina2022}} \\
&  &  including harm to individuals or the environment &  \\
\cline{0-0}\cline{3-4}
\multirow{2}{*}{Obligation to report emissions} &  & Companies are obliged to disclose the greenhouse gas emissions they produce to & \multirow{2}{*}{\cite{Bronnimann2019,Hanski2019,Casalino2020,Nabipour2020,Karasan2021}}\\
 &  &comply with general regulations or identify any excess emissions & \\
 \hline
Shift to less carbon-intensive  & \multirow{8}{*}{Technological} & \multirow{2}{*}{Use of fuels with lower greenhouse gas emission factors for thermal energy generation} & \multirow{2}{*}{\cite{Bronnimann2019,Hanski2019,Djenadic2022}}\\
production or consumption patterns &  &  & \\
\cline{0-0}\cline{3-4}
Technological progress in renewable   &  & Investments in products, processes, or services aimed at reducing carbon footprint  & \multirow{2}{*}{\cite{Bronnimann2019,Hanski2019,Aliyev2022,Djenadic2022}}\\
energies and energy efficiency &  & and improving environmental conditions, but that do not meet expectations & \\
\cline{0-0}\cline{3-4}
Technological change &  & New technological advancements that enable improved outcomes & \multirow{2}{*}{\cite{Huang2018,Sakhel2017,Katipoğlu2022,Djenadic2022,AkGul2019}}\\
 (development of new technology) &  & in the company's energy processes & \\
\cline{0-0}\cline{3-4}
Failed investments in new  &  & Development of new technologies with a less harmful impact on climate or the  & \multirow{2}{*}{\cite{Ak2019,Studenikin2021,Gallab2019,Bacanli2009,Hajiaghaei-Keshteli2023,Beiranvand2023}}\\
 technologies to reduce emissions &  & environment, rendering them obsolete or uncompetitive (stranded assets) & \\
\hline
Change in the demand for & \multirow{8}{*}{Market} & Changes in the demand for products and services & \multirow{2}{*}{\cite{Bronnimann2019,Huang2018,Abed2023,Ak2019,Bacanli2009,Nabipour2020,Karasan2021,Katipoğlu2022,Aydin2021}}\\
 products and services &  & which driven by concerns about climate change & \\
\cline{0-0}\cline{3-4}
Raw materials and supplies &  & Changes in prices, demand, volatility, and other aspects related  & \multirow{2}{*}{\cite{Bronnimann2019,Hanski2019,Abed2023,Karasan2021}}\\
(price volatility and availability) &  & to climate change that impact the supply of raw materials &\\
\cline{0-0}\cline{3-4}
Stakeholder concerns &  & Concern among market stakeholders and/or other& \multirow{2}{*}{\cite{Bronnimann2019,Hanski2019,Huang2018,Ak2019,Aliyev2022}}\\
on climate change &  & affected government and social groups, creating uncertainty & \\
\cline{0-0}\cline{3-4}
Poor adaptation to change  &  & Limited adaptability of business models to the changing  & \multirow{2}{*}{\cite{Bronnimann2019,Hanski2019}}\\
in customers' behavior &  & needs, desires, and customer behaviors regarding climate change& \\
\hline
Changes in customers preferences & \multirow{7}{*}{Reputational} & Reputation risk from loss or change in preference for a company's product & \cite{Bronnimann2019,Hanski2019,Sakhel2017,Abed2023,Tripathy2018,Ak2019,AkGul2019,Senol2021,Seddik2023}\\
\cline{0-0}\cline{3-4}
Increasing pressure from  &  & Pressures exerted by non-governmental organisations on companies' actions 	& \multirow{2}{*}{\cite{Nabipour2020}} \\
non-governmental organisations &  & with environmental impact, generate media attention and cause reputational harm	&  \\
\cline{0-0}\cline{3-4}
Negative news and comments &  & News on environmental responsibility enhance shareholder value,   &	\multirow{2}{*}{\cite{Bronnimann2019,Hanski2019,Karasan2021,Hajiaghaei-Keshteli2023}}\\
 Information about the company &  & while negative ones can have a deleterious impact on it.&	\\
\cline{0-0}\cline{3-4}
Changes in market sentiment due  & & Shifts in sentiment due to awareness of climate issues that  & \multirow{2}{*}
{\cite{Aliyev2022}} \\
to potential future climate risks &  & the future may hold if we cannot react promptly &  \\
\hline\hline
\end{tabular}
\end{adjustbox}
\caption{Climate transition variables selected}
\label{tab:table_3}
\end{table}

\subsection{Calculation of criteria weights}

First, the TOPSIS method was used to perform calculations to determine the scores according to the individual importance assigned by the experts and the ranking of the criteria determined by the climate transition risk assessment. The experts used a scale from 1 to 5 to rate each criterion, where 1 means "not at all important" and 5 means "very important". Table \ref{tab:table_4} shows the weights of the criteria as a result of applying the TOPSIS method. The modelling of the TOPSIS method equations was coded by using the Python programming language to facilitate the calculations involved.

\begin{table}[ht!]
    \centering
    \begin{adjustbox}{ width = 0.3\linewidth}
    \begin{tabular}{c|cccc}
    \hline\hline
\textbf{Criteria} & \textbf{Weights} \\ 
\hline
Vulnerability & 0.211 \\
Resilience & 0.183 \\
Exposure & 0.204 \\
Likelihood & 0.151 \\
Impact & 0.250 \\
\hline\hline
\end{tabular}
\end{adjustbox}
\caption{Ranking criteria}
\label{tab:table_4}
\end{table}

According to the weights assigned to the criteria, likelihood is the criterion with the lowest weight. It is important to note that both probability and impact are criteria used to determine the level of risk in multiple variables and in most risk assessment contexts. However, according to \cite{Bingler2020}, the impact is the result of the interaction between vulnerability, resilience, and exposure. This means that the results of the analysis can be interpreted as consistent from a theoretical point of view.

\subsection{MCDM method comparison}

The similarity or difference between different MCDM methods can be determined using various metrics, such as correlation coefficients. By examining the ranking table of the 10 selected methods, those that classify alternatives similarly are COPRAS, SAW, MARCOS and WSM. On the other hand, some methods, such as VIKOR, PROMETHEE, and ELECTRE, appear to classify these variables differently from the others. However, the choice of the most similar or dissimilar methods will depend on the method chosen to measure this similarity.

The RM2 risk is consistently high in most of the methods evaluated, as it is always in the first and second positions. On the other hand, the alternative or variable Rrep4 for reputational climate risk consistently ranks last in all methodologies.

\begin{table}[ht!]
    \centering
    \begin{adjustbox}{ width = \linewidth}
    \begin{tabular}{c|c|cccccccccc}
    \hline\hline
\multicolumn{2}{c|}{\textbf{Risk}}  & \multirow{2}{*}{\textbf{TOPSIS}} & \multirow{2}{*}{\textbf{COPRAS}} & \multirow{2}{*}{\textbf{BORDA}} & \multirow{2}{*}{\textbf{SAW}} & \multirow{2}{*}{\textbf{ELECTRE}} & \multirow{2}{*}{\textbf{VIKOR}} & \multirow{2}{*}{\textbf{MARCOS}} & \multirow{2}{*}{\textbf{PROMETHEE}} & \multirow{2}{*}{\textbf{WSM}} & \multirow{2}{*}{\textbf{CODAS}}\\ 
 \multicolumn{1}{c}{\textbf{Descritpion}}& \textbf{Code}  &  &  & &  &  &  & &  & & \\ 
\hline
\multirow{2}{*}{Cap and Trade Schemes} & \multirow{2}{*}{Rreg1} & \multirow{2}{*}6  &	\multirow{2}{*} 7 & \multirow{2}{*}7 & \multirow{2}{*}7 & \multirow{2}{*}8 & \multirow{2}{*} {10} & \multirow{2}{*}7 & \multirow{2}{*}6 & \multirow{2}{*}7 & \multirow{2}{*}7\\
 &  & & & & & & & & & &\\
\hline
\multirow{2}{*}{Carbon tax increase} & \multirow{2}{*}{Rreg2} &  \multirow{2}{*} 1 & \multirow{2}{*} 2 & \multirow{2}{*} 1 & \multirow{2}{*}2 & \multirow{2}{*}2 & \multirow{2}{*}2 & \multirow{2}{*}2 & \multirow{2}{*}1 & \multirow{2}{*}2 & \multirow{2}{*}2\\
&  & & & & & & & & & & \\
\hline
\multirow{2}{*}{Climate change-related litigation} & \multirow{2}{*}{Rreg3} & \multirow{2}{*} {12}  & \multirow{2}{*} {12} & \multirow{2}{*}{13} & \multirow{2}{*}{12} & \multirow{2}{*}{15} & \multirow{2}{*}7 & \multirow{2}{*}{12} & \multirow{2}{*}{13} & \multirow{2}{*}{12} & \multirow{2}{*}{12}  \\
&  & & & & & & & & & &  \\
\hline
\multirow{2}{*}{Obligation to report emissions} & \multirow{2}{*}{Rreg4} &  \multirow{2}{*} {10} & \multirow{2}{*} {10} & \multirow{2}{*}{10} & \multirow{2}{*}{10} & \multirow{2}{*}{10} & \multirow{2}{*}{11} & \multirow{2}{*}{10} & \multirow{2}{*}{11} & \multirow{2}{*}{10} & \multirow{2}{*}{10}\\
 &  & & & & & & & & & & \\
 \hline
Shift to less carbon-intensive  & \multirow{2}{*}{RT1} & \multirow{2}{*} 5 & \multirow{2}{*}5 & \multirow{2}{*}5 & \multirow{2}{*}5 & \multirow{2}{*}4 & \multirow{2}{*}5 & \multirow{2}{*}5 & \multirow{2}{*}5 & \multirow{2}{*}5 & \multirow{2}{*}4 \\
production or consumption patterns &  & & & & & & & & & & \\
\hline
Technological progress in renewable   & \multirow{2}{*}{RT2} & \multirow{2}{*} 7  & \multirow{2}{*}6 & \multirow{2}{*} 6 & \multirow{2}{*} 6 & \multirow{2}{*} 7 & \multirow{2}{*} 6 & \multirow{2}{*} 6 & \multirow{2}{*} 7 & \multirow{2}{*} 6 & \multirow{2}{*} 6\\
energies and energy efficiency &  & & & & & & & & & & \\
\hline
Technological change & \multirow{2}{*}{RT3} & \multirow{2}{*} 3 & \multirow{2}{*}3 & \multirow{2}{*} 3 & \multirow{2}{*} 3 & \multirow{2}{*} 3 & \multirow{2}{*} 1 & \multirow{2}{*} 3 & \multirow{2}{*} 3 & \multirow{2}{*} 3 & \multirow{2}{*} 3 \\
 (development of new technology) &  & & & & & & & & & & \\
\hline
Failed investments in new  & \multirow{2}{*}{RT4} & \multirow{2}{*} {13}  & \multirow{2}{*}{13} & \multirow{2}{*} {12} & \multirow{2}{*} {13} & \multirow{2}{*} {12} & \multirow{2}{*} {14} & \multirow{2}{*} {13} & \multirow{2}{*} {12} & \multirow{2}{*} {13} & \multirow{2}{*} {13}\\
 technologies to reduce emissions &  & & & & & & & & & & \\
\hline
Change in the demand for & \multirow{2}{*}{RM1} & \multirow{2}{*} {14} & \multirow{2}{*}{15} & \multirow{2}{*} {14} & \multirow{2}{*} {15} & \multirow{2}{*} {13} & \multirow{2}{*} {15} & \multirow{2}{*} {15} & \multirow{2}{*} {14} & \multirow{2}{*} {15} & \multirow{2}{*} {14}\\
 products and services &  & & & & & & & & & & \\
\hline
Raw materials and supplies & \multirow{2}{*}{RM2} & \multirow{2}{*} 2  & \multirow{2}{*} 1 & \multirow{2}{*} 2 & \multirow{2}{*} 1 & \multirow{2}{*} 1 & \multirow{2}{*} 4 & \multirow{2}{*} 1 & \multirow{2}{*} 2 & \multirow{2}{*} 1 & \multirow{2}{*} 1\\
(price volatility and availability) &  & & & & & & & & & & \\
\hline
Stakeholder concerns & \multirow{2}{*}{RM3} & \multirow{2}{*} 4 & \multirow{2}{*}4 & \multirow{2}{*} 4 & \multirow{2}{*} 4 & \multirow{2}{*} 5 & \multirow{2}{*} {13} & \multirow{2}{*} 4 & \multirow{2}{*} 4 & \multirow{2}{*} 4 & \multirow{2}{*} 5\\
on climate change &  & & & & & & & & & & \\
\hline
Poor adaptation to change  & \multirow{2}{*}{RM4} & \multirow{2}{*} {15}  & \multirow{2}{*}{14} & \multirow{2}{*} {15} & \multirow{2}{*} {14} & \multirow{2}{*} {14} & \multirow{2}{*} 3 & \multirow{2}{*} {14} & \multirow{2}{*} {15} & \multirow{2}{*} {14} & \multirow{2}{*} {15}\\
in customers's behaviour &  & & & & & & & & & & \\
\hline
\multirow{2}{*}{Changes in customers preferences} & \multirow{2}{*}{Rrep1} &  \multirow{2}{*} {11} & \multirow{2}{*} {11} & \multirow{2}{*} {11} & \multirow{2}{*} {11} & \multirow{2}{*} {11} & \multirow{2}{*} 8 & \multirow{2}{*} {11} & \multirow{2}{*} {10} & \multirow{2}{*} {11} & \multirow{2}{*} {11}\\
 &  & & & & & & & & & & \\
\hline
Increasing pressure from  & \multirow{2}{*}{Rrep2} & \multirow{2}{*} 8 	& \multirow{2}{*}8 & \multirow{2}{*} 8 & \multirow{2}{*} 8 & \multirow{2}{*} 6 & \multirow{2}{*} {12} & \multirow{2}{*} 9 & \multirow{2}{*} 8 & \multirow{2}{*} 8 & \multirow{2}{*} 9\\
non-governmental organisations &  & & & & & & & & & & \\
\hline
Negative news and comments & \multirow{2}{*}{Rrep3} & \multirow{2}{*} 9   &	\multirow{2}{*}9 & \multirow{2}{*} 9 & \multirow{2}{*} 9 & \multirow{2}{*} 9 & \multirow{2}{*} 9 & \multirow{2}{*} 9 & \multirow{2}{*} 9 & \multirow{2}{*} 9 & \multirow{2}{*} 8\\
 Information about the company &  & & & & & & & & & & \\
\hline
Changes in market sentiment due  & \multirow{2}{*}{Rrep4} & \multirow{2}{*} {16} & \multirow{2}{*} {16} & \multirow{2}{*} {16} & \multirow{2}{*} {16} & \multirow{2}{*} {16} & \multirow{2}{*} {16} & \multirow{2}{*} {16} & \multirow{2}{*} {16} & \multirow{2}{*} {16} & \multirow{2}{*} {16}\\
to potential future climate risks &  & & & & & & & & & &  \\
\hline\hline
\end{tabular}
\end{adjustbox}
\caption{Ranking of MCDM methods}
\label{tab:table_5}
\end{table}

There are clear distinctions between the techniques used for analysis. For example, the RM2 risk is generally as the most appropriate option by most methods, by most methods, with the exception of TOPSIS, BORDA, VIKOR and PROMETHEE. These discrepancies may be due to variation in decision criteria, weighting methods \cite{Hoang2022} or even the decision matrix, which differs from method to method. MCDM methods have different mathematical approaches, decision making principles, focus on different aspects, and also have different advantages and disadvantages. Some methods take into account the relative importance (weights) of the criteria, while others do not. This can result in different methods to produce different rankings or evaluations of alternatives and is one of the main causes of inconsistent results.

The complexity of real-world problems, the accuracy of the data, the possible relationships between criteria, the variation of trade-off indices or different types of scales can all influence the results. In this scenario of the climate transition risk analysis scenario, there is consensus among most of the methods. Looking at the results, it is clear that the risk RM2 is ranked number 1 by most methods (6/10), the risk Reg2 is ranked number 2 (7/10 methods), and the risk RT3 is ranked number 3 (9/10 methods); see Table \ref{tab:table_5}.

However, it is important to analyse the correlation between the MCDM methods; the following figure shows the correlation coefficients of the different methods evaluated.

\begin{figure}[htbp]
\centerline{\includegraphics[width=15cm]{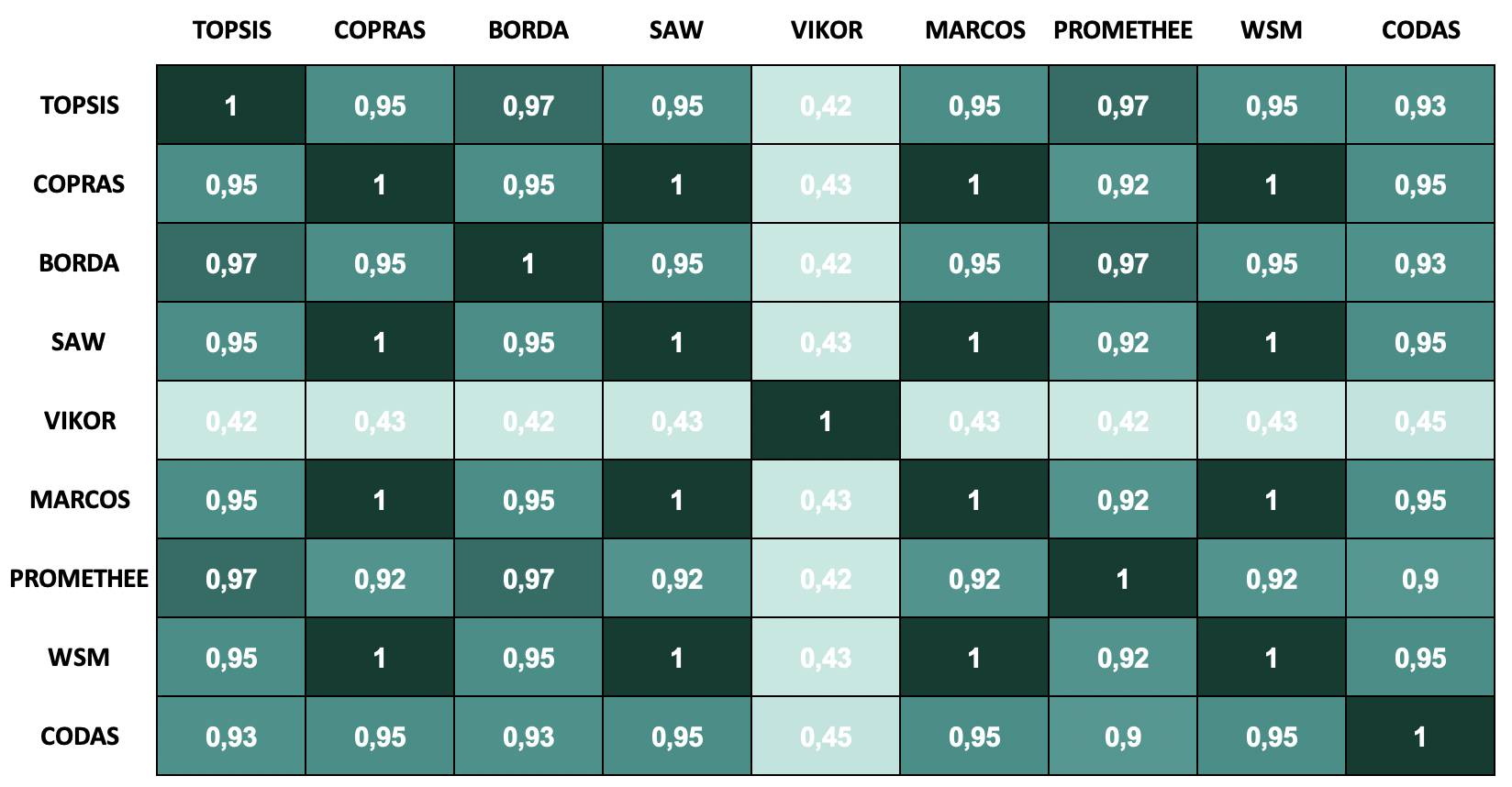}}
\caption{Correlation matrix}
\label{fig:Fig_6}
\end{figure}

Examining the correlation of weights between different MCDM techniques provides information to understand the similarities and differences between MCDM methods. Each method has its own algorithm and analytical approach for assigning weights to each decision criterion, which shows the relative importance of each criterion. Thus, the correlation of weights between different methods can be a useful indicator of the degree of agreement between these methods in prioritising decision criteria.

The rankings produced by the different MCDM methods show a high correlation, which implies that the options are considered similar due to their characteristics. This could be an indication of the reliability and robustness of the results, as different methods with different assumptions and procedures produce similar results.

It is important to keep in mind that, although a strong correlation may be reassuring, no MCDM approach should be considered the "correct" one. There may be elements that affect the overall score of an option that are not taken into account or are not fully taken into account by the different MCDM techniques. Therefore, it is still necessary to understand the assumptions of each method and interpret the results accordingly. For example, some techniques may be more appropriate for certain types of decision contexts or depend on certain types of data availability or quality.

Considering that MCDM methods only allow the generation of a ranking of alternatives, there is a limitation in the construction of risk matrices, as the results are not classified by risk level (low, medium, high, and critical), as shown in the risk matrix in Figure \ref{fig:matrix}. However, based on the results of the analysis of all MCDM methods, it can be determined that the risks RM2, Reg2, RT3, and RM3 are the most critical risks for the organisation of this sector. On the other hand, risks RT1, RT2 and Reg1 could be classified at the high level of the risk matrix, followed by risks Rrep2, Rrep3, Reg4, Rrep1, Rep3, RT4, RM4, RM1 and Rrep4, at the medium level of the risk matrix.

Comparing the results generated by the MCDM and Fuzzy Logic methods, we can see that RM2 and Reg2 are the most critical risks according to the results. However, not all risks were in the same order; in fact, only the risks RT4 and Rrep4 coincided, which were placed at positions 11 and 16, respectively, in both techniques used. However, the advantages of using methods such as Fuzzy Logic are the possibility not only to generate a ranking of alternatives, risks in this case, but also to classify them in different risk scales, as sought in this study. Furthermore, the results obtained with Fuzzy Logic coincide to a certain extent with those reported in the literature and by different organisations that analyse and monitor climate transition risks. The results obtained with the Fuzzy Logic method, which was selected for this study, are analysed below.

\subsubsection{Fuzzy-MCDM methods}

Many studies have combined Fuzzy Logic with various MCDM techniques, using the latter for the prioritisation of a set of alternatives. Determining the relative importance of various criteria in MCDM problems implies a high degree of subjectivity derived from the personal preferences of decision makers. However, the values of linguistic variables in expert judgment are often vague, representing interval values rather than fixed values; therefore, fuzzy MCDM methods often deal with this uncertain, imprecise, and vague information in decision-making problems \cite{Fu2021}. Among the different hybrid methods, we find combinations of Fuzzy with TOPSIS, AHP, ELECTRE, VIKOR, COPRAS, and VSM, among others.

Usually, when applying the Fuzzy-TOPSIS method (FT), the membership function is constructed first in the Fuzzy method, then the calculations of the fuzzy risk priority numbers are performed by multiplying the probability and impact membership functions, and finally these fuzzy risk level results are prioritised using the TOPSIS method \cite{Khodadadi2021}.

Table \ref{tab:table_6} shows the results of applying the FT method and the conventional TOPSIS method, showing a difference in the classification results. On the one hand, the risk prioritised by the FT method is market risk 3 (RM3), while in the TOPSIS method it is regulatory risk 2 (Rreg2). On the other hand, the second most important risk for each FT and TOPSIS method was the regulatory risk 2 (Rreg2) and market risk 2 (RM2), respectively. Overall, there are some similarities in the results, but given the differences in the application of the methods, there are obvious discrepancies.

In fact, the results of the Fuzzy-TOPSIS method differ to some extent from the results of the previously tested MCDM methods and the modified fuzzy method in this study. In this scenario, the results of the TOPSIS method are more consistent with the results of the previous section and with the individual responses of the consulted experts; for this reason, TOPSIS is selected as a complement to generate the weights in the fuzzy method proposed in this study. The final results obtained are discussed in the following subsection.

\begin{table}[ht]
    \centering
\adjustbox{width=0.4\textwidth}{
    \begin{tabular}{c|cc} \hline \hline
         \textbf {Risk} &  \multicolumn{2}{c}{\textbf {Rank}}\\ 
         \textbf {Code} &  \textbf {Fuzzy-TOPSIS}& \textbf{TOPSIS}\\ \hline 
         Rreg1& 8 &  6\\ \hline 
         Rreg2& 2 &  1\\ \hline 
          Rreg3& 15 &  12\\ \hline 
         Rreg4& 14 &  10\\ \hline
 RT1& 16 & 5\\\hline
 RT2& 7 & 7\\\hline
 RT3& 6 & 3\\\hline
 RT4& 12 & 13\\\hline
 RM1& 11 & 14\\\hline
 RM2& 5 & 2\\\hline
 RM3& 1 & 4\\\hline
 RM4& 10 & 15\\\hline
 Rrep1& 9 & 11\\\hline
 Rrep2& 4 & 8\\\hline
 Rrep3& 3 & 9\\\hline
 Rrep4& 13 & 16\\\hline \hline
    \end{tabular}
}
    \caption{Fuzzy-TOPSIS vs. TOPSIS methods}
    \label{tab:table_6}
\end{table}

\subsection{Fuzzy risk matrix analysis}

In this study, we used the centre of the area, also known as centroid defuzzification, which produces as a result a single numerical value corresponding to the fuzzy sets. Table \ref{tab:table_7} reports these numerical values for each of the 16 risks evaluated. Note that these results are obtained by processing the data from the procedure described in Section \ref{sec:datacollection} with the proposed TOPSIS approach, which allowed us to quantify the probability and impact criteria evaluated by the experts taking out questioonare. The TOPSIS method was selected as an alternative for determining the weights because, when analysing the results of the experts' opinions, it is clear that the opinions are very much in line with the weighting carried out by the TOPSIS method. These results reflect the concrete data obtained after applying the fuzzy inference model in the analysis of identified risks.

\begin{table}[ht]
    \centering
\adjustbox{width=0.8\textwidth}{
    \begin{tabular}{c|c|c|c|c|c|c} \hline \hline
         \textbf {Risk} &  \multicolumn{2}{c|}{\textbf {Likelihood}}&  \multicolumn{2}{c|}{\textbf {Impact}} & \multicolumn{2}{c}{\textbf {Risk level}}\\
         \textbf {Code} & \multicolumn{1}{c}{\bf{\%}} & \multicolumn{1}{c|}{\textbf{Linguistic}} &   \multicolumn{1}{c}{\bf{\%}} & \multicolumn{1}{c|}{\textbf{Linguistic}} &  \multicolumn{1}{c}{\bf{\%}} & \multicolumn{1}{c}{\textbf{Linguistic}}\\\hline 
         Rreg1&  0.3929&  Unlikely&  0.9371& High & 0.6000&High\\ \hline 
         Rreg2&  0.7321&  Likely&  0.9720&  High impact & 0.8644&Critical\\ \hline 
          Rreg3&  0.2000 &  Very unlikely&  0.8252&  Medium-high& 0.4000&Medium\\ \hline 
         Rreg4&  0.6071&   Medium&  0.7552&  Medium-high& 0.6000&High\\ \hline
 RT1& 0.8393& Likely& 0.9720& High & 0.8519&Critical\\\hline
 RT2& 0.4643 &  Medium& 0.9091& High & 0.6000&High\\\hline
 RT3& 0.6786& Likely& 0.9441& High & 0.7883&Critical\\\hline
 RT4& 0.2000 & Very unlikely& 0.8252& Medium-high& 0.4000&Medium\\\hline
 RM1& 0.2321& Very unlikely& 0.8392& Medium-high& 0.4696&Medium\\\hline
 RM2& 1.0000& Very unlikely& 1.0000& High & 0.8667&Critical\\\hline
 RM3& 0.4286 & Unlikely& 0.8741& High impact & 0.5202&High\\\hline
 RM4& 0.2000 & Very unlikely& 0.8042& Medium-high& 0.4000&Medium\\\hline
 Rrep1& 0.2143& Very unlikely& 0.8531& Medium-high& 0.4461&Medium\\\hline
 Rrep2& 0.4286 & Unlikely& 0.8182& Medium-high& 0.4559&Medium\\\hline
 Rrep3& 0.3750& Unlikely& 0.7832& Medium-high& 0.4000&Medium\\\hline
 Rrep4& 0.2000 & Very unlikely& 0.6853& Medium-high& 0.3797&Medium\\\hline \hline
    \end{tabular}
}
    \caption{Crisp risk values}
    \label{tab:table_7}
\end{table}

\begin{figure}[htbp]
\centering
\setlength{\tabcolsep}{1pt}
\begin{tabular}{>{\footnotesize}r*5{C{c}}}
    & head=\textbf{Low impact} & head=\textbf{Low-Medium impact} & head=\textbf{Medium impact} & head=\textbf{Medium-High impact\,} & head=\textbf{High impact} \\[0.9em]
    \multicolumn{1}{l}{\multirow{1.5}{*}{\textbf{Very likely}}} &   g &   y  &  o  &    r & r \\[0.9em]
\multicolumn{1}{l}{\multirow{1.5}{*}{\textbf{Likely}}} & g &  y  &   y & o   &   r\\[0.9em]
  \multicolumn{1}{l}{\multirow{1.5}{*}{\textbf{Medium likelyhood}}} & g & g    & y    & o       & o  \\[0.9em]
   \multicolumn{1}{l}{\multirow{1.5}{*}{\textbf{Unlikely}}} &  g & g     & y             & y&  o \\[0.9em]
    \multicolumn{1}{l}{\multirow{1.5}{*}{\textbf{Very unlikely}}} & g& g  & g    & y   & y    
\end{tabular}
\begin{tikzpicture}[overlay, remember picture]
\node[] at (-1.06,0.4) {\tiny \textbf{RM2}};
\node[] at (-1.06,-0.3) {\tiny \textbf{Rreg2, RT3,}};
\node[] at (-1.06,-0.5) {\tiny \textbf{RT1}};
\node[] at (-1.06,-1.15) {\tiny \textbf{RT2}};
\node[] at (-1.06,-2.05) {\tiny \textbf{Rreg1, RM3}};
\node[] at (-2.85,-2.05) {\tiny \textbf{Rrep2, Rreg3}};
\node[] at (-2.85,-2.65) {\tiny \textbf{Rrep1, Rreg3, }};
\node[] at (-2.85,-2.85) {\tiny \textbf{Rrep4,RT4,}};
\node[] at (-2.85,-3.05) {\tiny \textbf{RRM1, RM4}};
  \end{tikzpicture}
\caption{Climate transition risk matrix {\textbf{Green}} is used to indicate \textit{low risk}, {\textbf{yellow}} to indicate \textit{medium risk}, {\textbf{orange}} to indicate \textit{high risk} and {\textbf{red}} to indicate \textit{critical risk}.}
\label{fig:matrix_final}
\end{figure}
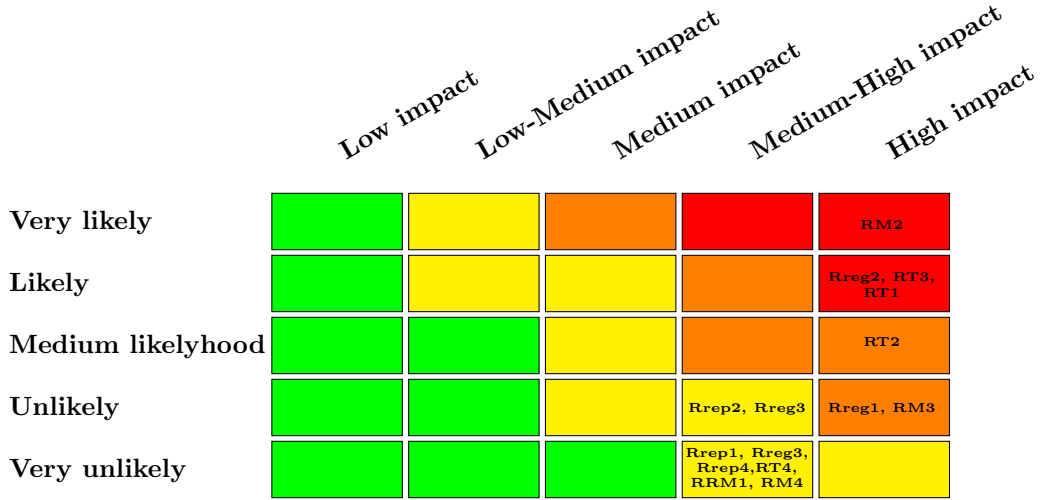

The above values were arranged in descending order to identify the most dominant risks, which were represented in the risk matrix shown in Figure \ref{fig:matrix_final}. When applying the fuzzy logic model and comparing it with the TOPSIS model, it was determined that the most critical and priority risks, in descending order, were RM2, RT3, Rreg2, and RT1, as can be seen in the red highlighted section in the matrix. These risks have a high probability of occurrence and a significant negative impact on the organisation, which is consistent with what has been reported in the specialised literature.

In terms of the most critical risk according to the results of experts' opinions, changes in prices, demand, volatility, and other aspects related to climate change will affect the availability of raw materials and inputs for companies (RM2) \cite{Demaria2021} \cite{Gasbarro2017}, generating uncertainty and increased costs of these supplies, increasing costs and production capacity \cite{Demaria2021}, due to changes in prices of certain inputs such as energy, water \cite{Campiglio2019} and fuels \cite{Semieniuk2021}. However, none of the methodologies we have evaluated so far considers the market approach from this perspective, as it is associated with the misperception of market climate risk as a long-term risk \cite{Jingyan2020}.

In relation to the increase in the carbon tax (Rreg2), which is a regulatory strategy that seeks to reduce GHG emissions by companies, studies such as \cite{Bankinternational2021} establish that one of the impacts of this risk is a reduction in operating margin due to a higher cost of CO2 emissions; a reduction in the prices of investments, shares \cite{Campiglio2021} and few profits, and higher interest rates, which in turn could limit access to financing and increase the cost of such financing \cite{Bankinternational2021}. Due to the additional payments resulting from the carbon tax, companies could face higher default costs and higher insurance premiums, as well as a possible impact on credit quality with high GHG emissions \cite{EuropeanSystemic2021}, \cite{Campiglio2021}.

It should be noted that most existing methodologies that estimate climate risk only employ the regulatory variable of carbon tax. However, according to \cite{Kolbel2021}[67], \cite{EuropeanSystemic2021} carbon emissions do not capture all aspects of climate transition risk; they are only a proxy of climate-related regulatory risks, given the multiple drivers and transmission channels, such as the potential for technological innovation and litigation \cite{EuropeanSystemic2021}.

However, in terms of the risks of RT1, RT2, and RT3, according to \cite{Campiglio2021} unexpected or very rapid changes in technological innovations related to climate, companies may also have economic and financial consequences, regardless of climate-related regulations. These innovations can turn companies' assets into obsolete or stranded assets (RT3), resulting in lower profit margins, higher operating costs, and financial destabilisation of the company \cite{Ahairwe2022}.

On the other hand, the transition to less carbon-intensive energy-based production patterns (RT1), which consists of the use of fuels with better GHG emission factors for the generation of thermal power in companies, would entail costs associated with the adoption of this new technology \cite{Demaria2021}. However, companies that rely on these types of processes and carbon-intensive technologies may become less competitive if they do not adopt these technological innovations \cite{Bankinternational2021}.

Complementing the analysis, the risks Rreg1, Rreg4, RT2 and RM3 are located in the orange zone (high risk) of the risk matrix. According to \cite{Gasbarro2017}, cap and trade schemes are schemes in which companies can trade their emissions (Rreg1) through certain established caps, which are more common in multinational companies. However, this can result in a high price for tradable emission permits \cite{cepal2017}, leading to higher operating costs, asset impairment, and early retirement of existing assets due to this change in regulation.

In relation to the Rreg4 risk, it is important to note that in Colombia the Superintendency of Finance issued External Circular 031, which regulates and encourages the disclosure of information on social, environmental and climate issues, under a financial materiality perspective. This regulation applies to companies listed on the Colombian Stock Exchange (BVC), that is, publicly traded companies.

However, this requirement for companies to report how much GHG emissions they produce and to determine whether they are in compliance with general regulations, or whether they are exceeding GHG emissions generation, may increase credit risk for banks and entities that do business with these companies. In addition, the information disclosed allows investors to make informed decisions about whether or not to invest in companies with certain levels of emissions \cite{Economist2015}.

However, the technological progress of renewable energy and energy efficiency processes (RT2), which consists of new technological developments that allow companies to achieve better results in their energy processes, will also generate direct transition costs with respect to renewable energy \cite{Ahairwe2022}, as well as higher operating costs.

In the context of stakeholder concerns about climate change (RM3), this type of risk could lead to investment aversion, resulting in losses on investments in companies, leading to losses \cite{Semieniuk2021}, as investors' awareness and expectations about climate change are increasing, they incorporate climate risk considerations into their investment decisions. Therefore, further research on the impact of corporate sentiment on climate change would help improve our understanding of this transmission channel \cite{Bankinternational2021}. Another impact that may be caused by climate change concerns is an increase in counterparty litigation, which generates other risks, such as reputational risks on certain environmental issues, which ultimately leads to increased costs and thus to reduced investment in unsustainable assets \cite{BankingAuthority2021}.

However, it is very important for companies to pay attention to changes in the environment, regardless of the risk typology, taking into account that the risks RM1, RM4, Rrep1, Rrep2, Rrep3, and Rrep4, have a medium risk rating. As the latter risks, which are related to the feelings and expectations of individuals and companies, are the least investigated \cite{Campiglio2021}.

\section{Conclusions and future work\label{sec:conclusions}}

It is important to note that MCDM methods are based on different principles and algorithms, as each was designed to solve specific problems. Some techniques weight the criteria differently, which can lead to different results for the same data set. On the other hand, if the data used are not at a similar level of analysis, it can distort the comparison. This may require the use of standardisation or normalisation techniques.

However, even if two MCDM techniques have a high correlation, it does not mean that one method influences the other; both could be influenced by external factors. It is important to note that correlation does not imply causation. Each method was designed for different problems or contexts, so it is important to understand the weaknesses and strengths of each method and to verify that it is appropriate for the problem being addressed in each investigation. Therefore, there is no one method that is more appropriate than another.

When comparing the results obtained by the MCDM and Fuzzy Logic techniques, it is possible to determine that RM2 and Reg2 are the most serious risks according to the results. Nevertheless, the advantage of using methods such as Fuzzy Logic is not only the ability to generate a ranking of options, in this case risks, but also to classify them into different risk levels, as was intended in this research.

The use of fuzzy logic, in combination with other decision-making techniques, can bring clarity and precision to the detailed study of the climate transition risks. The advantage of using fuzzy logic instead of other traditional qualitative approaches is the ability to transform vague and analogous information into unambiguous data that can be used to make informed and unambiguous decisions on various issues, such as climate transition risk management.

It can be concluded that technological risks are the most significant, followed by regulatory and market risks, for the economic sector analysed. All risks evaluated have a medium to high impact. The technological risks are particularly noteworthy due to their high probability of occurrence and their considerable impact. The risks RT1, RT2, and RT3 have a significant probability of occurrence and high impact, which makes them particularly critical.

Market and regulatory risks are the most significant in terms of risk management. In particular, RM2 is the most probable and has the most significant impact. Raw material price fluctuations are a common occurrence in the manufacturing sector, but are difficult to anticipate due to their unpredictable nature.

Reputational risks, although located in the lower right quadrant of the matrix, can have a significant impact and are not as widely studied by experts. Although their probability of occurrence is low, they can still have a significant impact, so it is essential to keep them under control.

From the above findings and analysis, it can be established that technological risks related to climate change are currently the most important for organisations in this sector, followed by regulatory and market risks, which are the key drivers for organisations at the financial level. Climate regulations seek to internalise carbon externalities as a key driver of risk, often through incentive-based regulation as reflected in carbon pricing. New technological developments help to reduce costs, sometimes incentivised by earlier climate regulation, thereby lowering the price of low-carbon technologies. Finally, consumer and market preferences will drive demand and prices. Given this demand-pull effect, market preferences can influence the pace of technology adoption and policy development, demonstrating the strong interaction between these three drivers of climate change.

However, it is essential for companies to pay attention to changes in the environment, regardless of the type of risk, taking into account that the risks RM1, RM4, Rrep1, Rrep2, Rrep3, and Rrep4 have a medium risk rating. This is because the latter risks, which relate to the feelings and expectations of individuals and companies, are the least researched.

Therefore, efforts to better address and manage climate change are expected to drive technological innovations that will enable the transition to low-carbon economies. This could make the most polluting technologies relatively more expensive if carbon taxes or other stricter regulations are introduced, as companies that rely on carbon-intensive technologies could become less competitive if they do not adopt new technologies.

Some results may differ from other analyses of climate transition risk assessment, taking into account that the experts assessed in this study belong to a company in the economic sector analysed, who know the reality of the company and who have knowledge associated with the risks, which is important in determining how exposed the company may be to this type of climate risk.

The severity of the threat that can affect an organisation depends on its vulnerability and resilience. Furthermore, the economic impact is the result of the interaction between exposure, vulnerability, and resilience. Therefore, an assessment of the financial consequences of climate transition risks seeks to quantify the economic impact on the underlying unit of analysis.

However, companies are expected to create a culture of identifying, assessing, and managing climate transition risks through this type of analytical tool. It should be noted that this tool can be used not only for the identification of purely climatic risks, but also for any type of traditional or emerging risk faced by companies, as well as the climate physical risks.

This research aimed to identify climate transition risks that could affect the organisations which belongs to the processed food sector. It is important to note that these specific risks were identified for a specific company, taking into account its context and realities. Exposure to a particular type of risk will depend largely on the sector and type of industry. Many risks can be added to this matrix as necessary for the analysis given the needs and specificities of each organisation.  

It is suggested that future research should focus on methods that can weigh vulnerability, exposure, and resilience on the same impact criterion, as input variables for the fuzzy logic model. This would allow experts to capture more information, similar to the MCDM model, since the literature suggests that there is a relationship between these four criteria.

On the other hand, it is suggested that for the application of this methodology in specific companies, a representative sample of the different areas of the organization should be used, since it is important that all areas without exception are involved in the process of risk assessment and subsequent evaluation. Considering as experts people with knowledge associated with the risks to which the business may be exposed.

\bibliographystyle{elsarticle-harv}
\bibliography{biblio}

\end{document}